# DeepFMEA – A Scalable Framework Harmonizing Process Expertise and Data-Driven PHM

Christoph Netsch[1], Till Schöpe[1], Benedikt Schindele[1], and Joyam Jayakumar[1]

[1]*Alpamayo Intelligent Quality Solutions, Sarnen, 6060, Switzerland*
*{netsch|schoepe|schindele|jayakumar}@alpamayo.ch*

## ABSTRACT

Machine Learning (ML) based prognostics and health monitoring (PHM) tools provide new opportunities for manufacturers to operate and maintain their equipment in a risk-optimized manner and utilize it more sustainably along its lifecycle. Yet, in most industrial settings, data is often limited in quantity, and its quality can be inconsistent – both critical for developing and operating reliable ML models. To bridge this gap in practice, successfully industrialized PHM tools rely on the introduction of domain expertise as a prior, to enable sufficiently accurate predictions, while enhancing their interpretability.

Thus, a key challenge while developing data-driven PHM tools involves translating the experience and process knowledge of maintenance personnel, development, and service engineers into a data structure. This structure must not only capture the vast diversity and variability of the expertise but also render this knowledge accessible for various data-driven algorithms.

These challenges result in data models that are heavily tailored towards a specific application and towards the failure modes the development team aims to detect and/or predict. The lack of a standardized modeling approach limits developments' extensibility to new failure modes, their transferability to new applications, and it inhibits the utilization of standard data management and MLOps tools, increasing the burden on the development team. In effect, high development and industrialization costs limit the economic utility of data-driven PHM tools to use cases with exceptionally high economic risks.

*DeepFMEA*, draws inspiration from the Failure Mode and Effects Analysis (FMEA) in its structured approach to the analysis of any technical system and the resulting standardized data model, while considering aspects that are crucial to capturing process and maintenance expertise in a way that is both intuitive to domain experts and the resulting information can be introduced as priors to ML algorithms. Our proposed framework promises a consistent use of best practices in data-driven modeling for PHM use cases while enhancing their interpretability, cost-effectiveness, and the scalability of their deployment.

## 1. INTRODUCTION

The widespread adoption of data-driven technologies has opened new horizons in various industrial applications, among which prognostics and health monitoring (PHM) stands out as a critical area of focus. PHM refers to the use of advanced analytical tools and techniques for the online monitoring of equipment, diagnostics of specific failure modes, and prognostics regarding the future performance or failure of machinery and equipment. By leveraging Machine Learning (ML) and other data-driven approaches, PHM aims to enable the adoption of condition-based and predictive maintenance strategies (Nunes et al., 2023). This promises to optimize maintenance schedules and minimize the risk resulting from the consequences of failures, such as operational downtime, reduced quality, or reduced energy efficiency, by enhancing overall system reliability and efficiency.

Original equipment manufacturers (OEMs), system integrators, maintenance, and repair organizations (MROs) and operators of large fleets are each well-positioned to develop and deploy PHM tools for a specific class of equipment. To varying degrees, they possess:

- access to in-operation data representative of the operational context, under which the equipment is operated, and access to maintenance and service data,
- an understanding of the equipment's design and its dominant failure modes and the experience, allowing for reasonable prior assumptions regarding consequences and failure rates,
- an understanding of the maintenance policies in place, and the requirements that determine how a PHM tool can be utilized to enable condition-based and predictive maintenance techniques.

OEMs and system integrators in particular are in a unique position to integrate these solutions natively into equipment at the point of production, promising close alignment between the equipment's design and the PHM tool. This includes access to the full range of available data from the system's intrinsic sensors and control signals and the ability to embed sensors to cover potential "gray spots" within the system. With the promises of improved asset reliability, longevity, and increased maintenance efficiency, they intend to create differentiating features for their products, their spare parts offering, competitive advantages for their aftersales services, or entirely new business models around their core product (Potthoff et al., 2023).

Reliability-centered maintenance (RCM) (Basson, 2019) and Total Productive Maintenance (TPM) (Wireman, 2004), among others provide robust theories on the circumstances, under which adopting a condition-based or predictive maintenance policy is both practical and cost-effective. However, given the imperfect nature of their predictions, the overall risk reduction potential of a PHM tool cannot be assessed without understanding how its deployment would impact the tasks realized by maintenance practitioners. This can be characterized by three key factors:

- the **accuracy** of that information, with respect to the rates of true detections, false detections, and missed detections, influencing savings related to avoided failure consequences and the additional costs of unnecessary diagnostic tasks.
- the **prediction horizon** characterized by the time between the detection of a potential failure and the observance of the corresponding functional failure, which corresponds to the P-F-interval for many failure patterns and determines the practicality of a proactive task.
- the **prescriptive value** of the information provided to the maintenance practitioner – a result of both the choice of algorithm and the interpretability of its output - influencing how directed the resulting task is.

Based on requirements stemming from a specific use case, and the skillsets and convictions of a particular team tasked with developing such systems, teams tend to make use of data-driven modeling techniques (classical statistical models and ML) or mechanistic modeling techniques to varying degrees. Mechanistic models are trusted for their explicit representation of domain knowledge and the interpretability of their outputs. However, with increasing complexity of a system and the uncertainty related to stochastic phenomena and incomplete knowledge of a system's operational context and environment, the gap between model and reality renders purely mechanistic models useless for many use cases (Eker et al., 2016; Hagemeyer et al., 2022)

Data-driven models, on the other hand, are able to capture complex relationships within data, given an incomplete representation of a specific asset's history and operational context, can frequently be refitted to new data, and, in theory, require no prior assumptions (Liao et al., 2016). However, the outputs created by purely data-driven models rarely provide the prescriptive value that condition-based and predictive maintenance demands for and enjoy less trust among maintenance practitioners, due to their lack of interpretability (Vollert et al, 2021). Additionally, in the context of PHM, the utility of purely data-driven modeling is severely limited, due to the variability and quality of the data upon which these systems rely (Luo et al., 2020, Nunes et al., 2023) and their limited availability - particularly in industrial applications, where data sharing collaborations between different organizations remain an exception (Trauth et al., 2020).

In practice, and with the exception of some examples of purely academically motivated research and proof-of-concept (PoC) implementations, the underlying models of PHM tools are often hybrid in nature (Luo et al., 2020). On one hand, hybrid modeling benefits from the adaptability and ability of data-driven models to learn from incomplete information. On the other hand, it leverages domain knowledge to enhance output interpretability and introduce priors into the model that compensate for a data-driven model's ability to generalize to operating contexts and failures, that are underrepresented in the available data. It is not surprising that the PHM community in particular has proposed noteworthy contributions to the field of hybrid ML, for instance the use of graph neural networks to learn representations of the semantic relationships of signals inherent to complex technical systems and to introduce known relationships as priors to a neural network (Battaglia et al., 2018, Zhao et al., 2020) or approaches designed to make connect available, yet incomplete or inaccurate physics-based models with ML-based models (Gassner et al., 2014, Arias Chao et al., 2022)

Our contribution does not challenge current fundamental contributions to hybrid modeling, nor does it extend the field's state-of-the-art. Rather, by proposing DeepFMEA as a standardized framework for the development and deployment of data-driven PHM tools our contribution recognizes the importance of these contributions, while strengthening the basis for their industrialization. To do so, in *section 2* we first derive requirements for a potential framework from current challenges. In *section 3,* we propose (3.1) a standardized data model allowing for a structured representation of domain knowledge in PHM use cases, (3.2) a non-comprehensive overview of methods to harmonize this information in hybrid data-driven modeling approaches, and (3.3) an overview of methods to enrich model outputs with prescriptive information, to enhance their actionability. In *section 4* we demonstrate an example of DeepFMEA usage in a practical use case based on monitoring and diagnostics for a hydraulic system (Helwig, 2015). Finally, *section 5* summarizes our findings from implementations of the framework in real-world PHM projects and discusses its current limitations and extension potentials.

## 2. CHALLENGES IN THE INDUSTRIALIZATION OF DATA-DRIVEN PROGNOSTICS AND HEALTH MONITORING TOOLS

In practice, the main effort of developing PHM solutions is not spend with model development. In order to deploy and operate data-driven PHM tools in a production environment, substantial development effort is introduced by the data and software engineering required to provide the following functions:

- **Data management** encompasses IoT-connectivity, data pipelines, centralized or decentralized databases, and data processing services (data health checks, preprocessing, inference, and postprocessing) that enable the tool's online operation.

- **Model monitoring & updates** implements feedback loops that automate the continuous evaluation of the tool's underlying model in production by relating its outputs to any available maintenance and operations data from a given asset. Furthermore, this encompasses the model training service, model registry, and decision logic required to adapt, update, and replace models in a production environment. Despite these capabilities falling into the category of MLOps (Kreuzberger et al., 2023) arguably, in PHM, they are not only required to operate ML-models, but data-driven models in general. This is due to the fact that datasets, PHM tools are initially developed with, rarely reflect the full range of operating contexts and failures, that a fleet of assets may experience in operation (Zio, 2022)

- **The process integration layer** includes reporting services and interfaces to enterprise software (i.e. ERP, CMMS) needed to make model outputs accessible to asset managers, field engineers, and maintenance professionals, i.e. as decision-support systems or automations. An additional function realized in the application layer of a PHM solution is to capture the maintenance and operations data required to realize the monitoring function.

The heavy reliance on domain knowledge, described in *section 1*, combined with the fact that PHM tools frequently rely on the ingestion and fusion of control level (PLC), process management level (SCADA), and/or management level (MES) data (Mantravi et al., 2022) commonly results in highly customized data models and data pipelines. While most requirements of the aforementioned functions can be abstracted to a level, where they are agnostic to a specific PHM use case, this high degree of customization introduces additional requirements for the data management, MLOps, and process integration layers that interface with them.

In effect, this inhibits the use of general-purpose software modules and increases the burden on the development team. Particularly the implementation of MLOps functions demands a highly specialized skillset (Nahar, 2022) which is rarely represented in PHM development teams.

Considering the prevalence of hybrid modeling approaches described in *section 1* and the challenges to their industrialization introduced in section 2, a framework for the development of data-driven PHM tools requires:

- a consistent representation of domain knowledge commonly relevant to the PHM use case, such as:
  - semantic relationships between sensors and the elements of a complex technical system they are localized to,
  - prior knowledge that facilitates the diagnostic and prognostic functions of a PHM tool with respect to specific failure modes, i.e. virtual sensors, detection rules, and degradation variables,
  - prescriptive information, i.e. recommended diagnostic, proactive, or reactive maintenance tasks, given process anomalies or a specific failure mode is detected,
- a data structure and systematic approach that connects in-operation data, maintenance data, and domain expertise via hybrid modeling approaches, yet maintaining the flexibility to integrate problem-specific approaches or advances in the state-of-the-art,
- a common approach towards how data-driven monitoring tools for the purposes of monitoring, diagnostics, and prognostics are managed in production, to facilitate the efficient use of general-purpose technology,
- a quantitative assessment of monitoring, diagnostics, and prognostics tools that quantifies their impact in terms of risk-reduction, considering the imperfect nature and the inherent uncertainty of data-driven model predictions.

## 3. CONCEPT

*DeepFMEA*, our proposed framework, does not draw its name being an implementation the Failure Modes and Effects Analysis (FMEA) (MIL-STD-1629A, 1980, Rausand et al., 2003) nor is it a data-driven extension to the FMEA process, as proposed by (Ervural & Ayaz, 2023). We draw inspiration from:

- the strong intersection between the data required by the FMEA and the information commonly used as domain expertise in the modeling process,

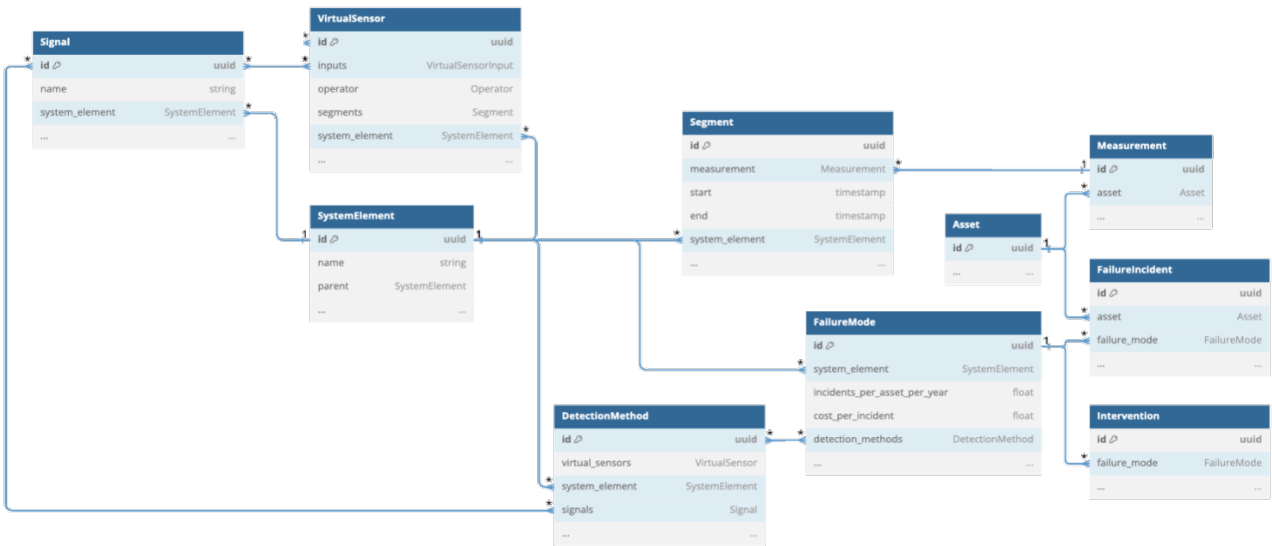

Figure 1: Simplified UML Diagram describing the data models used within the DeepFMEA Framework.

- the systematic process of capturing this data within an FMEA,
- the structured relational data model suggested by the analysis, which has been implemented in database schemas by numerous vendors of specialized FMEA-software.

This section briefly outlines the abstract data model implemented by *DeepFMEA* (3.1), before presenting a non-comprehensive collection of approaches to systematically utilize this information in the modeling process (3.2), enriching model outputs with prescriptive information (3.3), and assessing their impact on risk (3.4).

### 3.1. Data models

A *System Element* reflects the part of a system at which a failed state (the inability of a system to fulfill its intended function) can be localized at. Destructuring a complex system into system elements results in a hierarchical data structure, frequently resembling the subsystems, assemblies, components, or parts of its physical twin. The root system element corresponds to the system modelled in the PHM project.

An *Asset* corresponds to a physical entity of the root system element.

A *Signal* declares any class of in-operation data captured in a technical system, either from intrinsic sensors, control signals, or external sensors retrofitted to the system. A *Signal* references one or multiple *System Elements*.

A *Measurement* contains the time-series associated referencing a *Signal* of an *Asset* in a given time interval. Whenever a system's process is cyclic in nature, it is a good practice to define *Measurements* along the boundaries of cycles, since this already introduces strong normalization to the time-series simplifying their subsequent processing.

A *Segment* references the method and its parameters required to retrieve a given recurring pattern, i.e. corresponding to a specific step in a process or a procedure in the operation of a machine. It is model-agnostic. The underlying segmentation model could be a simple rule, a motif detection algorithm or even a neural network.

A *Virtual Sensor* defines the computation of a property of one or multiple *Measurements,* which is informed by domain knowledge. *Virtual Sensors* are commonly used in PHM projects due to their enhanced interpretability, expressiveness and better generalization properties compared to raw signals. An example of a *Virtual Sensor* in a shaft component is the "1X Frequency". Elevated levels are commonly (but not exclusively) associated with its misalignment. In analogy to a *Signal*, it references one or multiple *System Elements*. The virtual *Measurements* obtained can be scalar, vectorial or tensorial. The computation of a *Virtual Sensor* is stored as a graph of atomic *Operations*, that reference *Signals, Segments,* or other *Virtual Sensors. Section 3.2* introduces multiple concepts of how *Virtual Sensors* can be used in monitoring, diagnostics, and prognostics *Detection Methods*.

A *Failure Mode* defines one form in which a *System Element* can fail. In order to enable an assessment of risk and the risk-reduction potential corresponding to the deployment of a *Detection Method*, a *Failure Mode* includes properties that quantify the (economic) consequences of a *Failure Incident*. For the same purpose, due to the fact that *Failure Incidents* are rare in many PHM use cases, in practice, prior assumptions regarding the failure rate may have to be introduced as additional properties. Reasonable assumptions should be set in agreement between multiple domain experts.

An *Intervention* stores a prescriptive instruction directed to an asset's operator or maintainer, given occurrence of a specific *Failure Mode*. An *Intervention* can be a diagnostic, reactive, or proactive task and usually comes with a non-negligible cost.

A *Failure Incident* stores a detected or observed occurrence of a *Failure Mode* for a given *Asset*. In order to monitor the performance of a given *Detection Method*, its properties allow determining whether the *Failure Incident* was detected prior to its occurrence, remained undetected or was a false alarm.

A *Detection Method* references any monitoring, diagnostic, or prognostic model developed and deployed as part of the PHM tool. It carries references to both *Signals* and *Virtual Sensors*, that it relies on as inputs, and to the *System Elements* (in the case of monitoring) or *Failure Modes* (in the case of diagnostics and prognostics).

### 3.2. Systematic Use of Domain Knowledge in Detection Methods

Below, we present a selection of methods that illustrate how the data mentioned earlier can be systematically leveraged to develop data-driven PHM tools. Instead of categorizing these methods by specific algorithm classes, we organize our discussion around the typical progression of objectives in PHM projects. Projects often start with Proof of Concepts (PoCs) that focus primarily on monitoring the operation of assets by detecting deviations from normal processes. They then gradually extend the tool's capabilities to diagnose high-risk failure modes and, eventually, to forecast potential incidents. By examining the data requirements for each of these objectives, *DeepFMEA* ensures that a PHM tool can be expanded without significant alterations to its data and software architecture and the high-quality data, that is particularly important for diagnostics and prognostics, is collected from the onset of the project.

#### 3.2.1 Monitoring

Monitoring targets the detection of anomalous behavior in a process without pinpointing specific failure modes. employs anomaly detection algorithms that leverage readily available data from normal operations. Due to the complexity and high dimensionality of data from industrial equipment, direct application of data-driven models often leads to poor generalization. Moreover, it inhibits the ability to localize anomalies within a large system. Therefore, it is common practice to filter data and distill only the most pertinent information.

The data model defines the relationships between the *Detection Method*, the *System Element(s)* within its scope, and the associated *Signals*. It permits a preselection of data relevant to the monitoring objective while eliminating signals that do not enhance the monitoring tool's performance because they lack informational value about the focused *System Elements*.

Moreover, *Virtual Sensors*, associated with the System Elements under observation, can be input to anomaly detection models as health indicators to yield an anomaly score (or *attention index*). These indicators extract crucial information from the dense, noisy, and often overly detailed sensor data collected from real-world assets, ensuring the monitoring process is both efficient and effective.

#### 3.2.2 Diagnostics

Unlike monitoring, diagnostic tools offer operators and maintainers precise information that identifies a failure mode or several probable ones. Building on monitoring concepts, the data model's embedded domain expertise can be further utilized for diagnostic challenges.

Diagnostics typically takes the form of a classification problem, where *Failure Modes* defined in the data model represent the classes. Implementing this model in a database together with the data pipelines, that make the data accessible from its sources, results in a scalable data management system, aligning *Failure Incidents* with their respective *Failure Mode*. The continuous collection of failure data not only provides the data required to fit the diagnostic *Detection Method's* model or evaluate its performance metrics; it also creates the basis for automated monitoring, adaption, and updating of any data-driven models in a production environment.

Diagnostics use cases often face a notable lack of Failure Incidents, particularly for high-risk Failure Modes. To address this, PHM projects might rely on strong assumptions, like establishing explicit or inferred thresholds on *Virtual Sensor* values for a *Failure Mode's* occurrence, based on expert consensus.

Additionally, section 4 illustrates a simplified example showing how data on failure rates, consequences, and interventions can evaluate a *Detection Method's* risk reduction impact upon deployment.

#### 3.2.3 Prognostics

Prognostics aims to forecast specific Failure Modes by estimating the remaining useful life (RUL) of a System Element. Although data-driven models for prognostics have been a focal point in PHM research, their practical impact on industrial PHM tools remains minimal. The prominence of research in this area within parts of the PHM community might give the impression of their significant role, but in practice, the utility of RUL predictions is often limited. Despite their allure, they may be of limited value for many *Failure Modes* where monitoring or diagnostic tools can identify potential failures well in advance of the functional failure, allowing for their proactive management. Furthermore, for *Failure Modes* governed by stochastic events, is a futile endeavor.

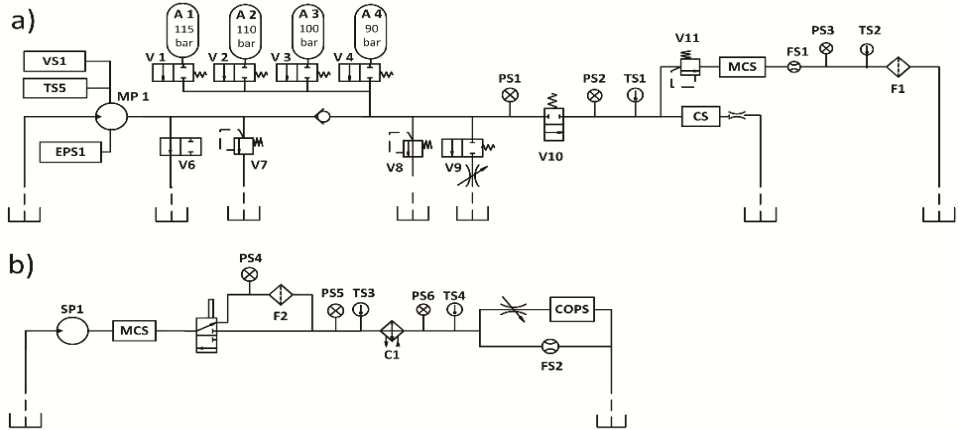

Figure 2. Hydraulic system testbench used to obtain training data for (Helwig, 2018).

Creating effective data-driven prognostics models requires extensive datasets, including both numerous *Failure Instances* and comprehensive in-operation data on the degradation trajectory. Such complete datasets are exceptionally rare in practice. However, a data management system implementing the *DeepFMEA* data model has the technical capability to compile datasets fulfilling these stringent criteria. When a *Failure Mode* is thoroughly understood, domain experts can identify *Virtual Sensors* as degradation variables, feeding into models that estimate RUL.

### 3.3. Enhancing the Prescriptive Value of PHM Outputs

Data-driven models often face criticism for their lack of interpretability. However, the rich contextualization *DeepFMEA* offers for a *Detection Method* can significantly enhance the model's outputs. By adding context data and prescriptive information before presenting results to operators and maintainers, the trustworthiness of a PHM tool is improved, making it a more valuable decision-support system.

When a deviation from the healthy process is detected, a monitoring tool can use the available information on the *System Element(s)* referenced by the *Detection Method* to provide suggestions for its origins and reduce the amount of time for its localization.

The presentation of context data can mirror the hierarchical structure of the physical system, allowing operators and maintainers to intuitively trace anomalies from a system-wide perspective down to specific components. This structured and visualized approach aligns with the natural process of investigating anomalies.

Furthermore, diagnostic, proactive, and reactive *Interventions* tied to particular *Failure Modes* can be directly communicated to the operator or maintainer. Alternatively, they can be seamlessly integrated into existing processes, automatically initiating the appropriate workflows. This integration facilitates a more effective and timely response to potential issues.

### 3.4. Impact as a decision-support System

As a general principle, a proactive task should be considered in the maintenance policy of an asset, when it effectively minimizes risk. The failure mode risk priority number ($RPN_{FM}$) involves rating the risk inherent to a failure mode given its probability of occurrence $P_{FM}$ in a given time interval, its severity $S_{FM}$, and the probability of its detection $D_{FM}$:

$$RPN_{FM} = P_{FM} * S_{FM} * D_{FM} \qquad (1)$$

It is common practice to use *RPN* to conduct a risk-based prioritization in the process of developing and reviewing a maintenance policy. For better readability, the subscript *FM* is omitted from here onwards. When *S* can be expressed in terms of costs when a failure is detected *CD* (i.e. cost of can derive a Quantitative Consequence Priority Number ($QCPN$) for the failure mode reflecting the expected cost per asset for in a given time interval:

$$QCPN = P * (D * CD + (1 - D) * CU) \qquad (2)$$

Once deployed as decision-support systems and automations the data-driven *Detection Methods* discussed in this paper play a similar role in the maintenance policy as conventional scheduled diagnostic tasks. They are feasible, when their utilization significantly increases the probability of detection of a failure $D^* \gg D$, however they introduce additional costs of the diagnostic task $CDI$. In comparison to conventional diagnostic tasks, the online *Detection Methods* discussed in this paper results in practically infinitesimally small detection intervals. In many cases this has a significant positive impact on $D^*$. On the other hand, the intrinsic uncertainty that generally characterizes data-driven models and the cost $C_{PHM}$ connected to the introduction and operation of the Detection Method itself need to be considered.

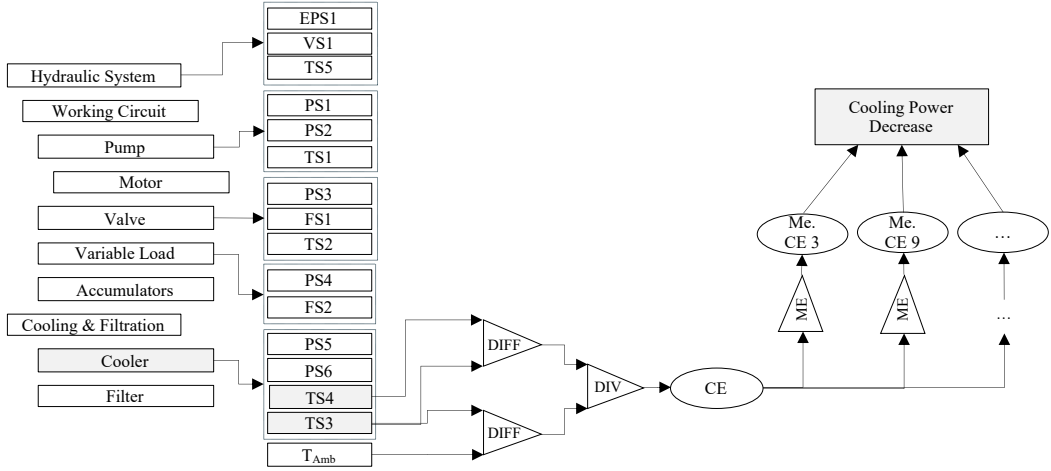

Figure 3. Visual description of the hydraulic systems including links used within the DeepFMEA data structure with the example of failure condition "Cooling Power Decreased".

proactive overhaul or replacement) or remains undetected $CU$ (i.e. cost of downtime, secondary damages, and repairs), we

Since most PHM tools are not deployed as automations, but as decision-support systems, we continue to factor in $C_{DI}$, with the difference that diagnostic tasks will not be executed on a scheduled basis, but in reaction to a detection.

Furthermore, we introduce the true positive rate ($TPR$), false positive rate ($FPR$), and false negative rate ($FNR$) as means to approximate $D^*$ using metrics we obtain by benchmarking a diagnostic PHM tool on historic *Failure Incidents*.

$$QCPN^* = P * (TPR * CD + FNR * CU + FPR * CDI) + C_{PHM} \quad (3)$$

We can then yield the expected impact of the deployment of the PHM tool ($\Delta QCPN_{PHM,FM}$) as the difference of (2) and (3):

$$\Delta QCPN_{PHM} = P * (D * CD + (1 - D) * CU) \quad (4)$$
$$- P * (TPR * CD + FNR * CU + FPR * CDI) + C_{PHM}$$

a positive value indicating the PHM tool is cost-effective in reducing the expected cost per asset for a given time interval.

## 4. REFERENCE USE CASE: HYDRAULIC SYSTEM DATASET

To demonstrate the use of the proposed *DeepFMEA* framework, we show its utilization with Helwig's hydraulic system dataset (Helwig, 2018). This dataset was acquired on a hydraulic test rig, where different degradation mechanisms and multiple failure modes were induced. The setup, consisting of a working circuit including a variable load and a cooling and filtration circuit, is depicted in Figure 1. The rig is equipped with pressure, temperature, vibration, and volume flow sensors at different components. We replicate the approach in our *DeepFMEA* framework.

We start by describing the physical machine as a tree-structure of *System Elements*, then continue to link all available *Signals* to their respective *System Elements* as indicated in Tables 1 & 2 respectively. The hierarchy permits aggregating *Signals*, while also considering child components. For instance, despite EPS1 and VS1 not referencing the exact same *System Element*, they can all be aggregated to the parent component "Pump".

*Segments* partition the data into time ranges that provide comparable classes of samples within the dataset (Table 3) based on the hydraulic rig's variable load.

Helwig further proposes computing scalar features and virtual sensors meaningful to the problem (Helwig, 2015). They are predominantly based on common thermo- and hydrodynamic KPIs, like cooling efficiency and power, and first principles, like the heat transfer equation. We split their computation into atomic mathematical operations that receive raw data from *Signals*, *Segments*, and the results of other *Virtual Sensor* computations as inputs and output a new

*Virtual Sensor* as indicated in Table 4, providing a means to represent complex virtual sensors in a uniform data structure.

Furthermore, we define known *Failure Modes*: Cooling power decrease, switching characteristic degradation of the valve, internal leakage of the pump and gas leakage of one of the accumulators. Establishing sensible references to the corresponding *System Elements* is straightforward in this example (Table 5). Labels of the original dataset are incorporated as *Failure Incidents* and *Interventions* respectively.

Table 1. Simplified System Element data-model.

| Name | Parent (FK) |
|---|---|
| Hydraulic-System | NULL |
| Working-Circuit | Hydraulic-System |
| Pump | Working-Circuit |
| Motor | Pump |
| Valve | Working-Circuit |
| Variable Load | Working-Circuit |
| Accumulators | Cooling & Filtration |
| Cooling & Filtration | Working-Circuit |
| Cooler | Cooling & Filtration |

Table 2: Simplified signal data-model.

| Name | System Element (FK) | Sampling-Rate |
|---|---|---|
| EPS1 | Motor | 100 Hz |
| VS1 | Pump | 1 Hz |
| PS1 | Valve | 100 Hz |
| … | | |
| TS3 | Cooler | 1 Hz |
| TS4 | Cooler | 1 Hz |

Table 3. Simplified Segment data-model.

| Name | Start | End |
|---|---|---|
| INT1 | 0.00s | 60.00s |
| … | | |
| INT13 | 50.01 | 60.00s |

Table 4. Simplified Virtual Sensor data-model.

| Name | Inputs | Operator |
|---|---|---|
| $\Delta T_{Cool}$ | (T3, INT1) (T4, INT1) | DIFF |
| $\Delta T_{Amb}$ | (T3, INT1) (TAmb, INT1) | DIFF |
| … | | |
| CE | $\Delta T_{Cool}$ $\Delta T_{Amb}$ | DIV |
| Median CE 3 | (CE, INT3) | ME |

Table 5. Simplified Failure Mode data model.

| Name | System Element (FK) |
|---|---|
| Cooling Power Decrease | Cooler |
| Internal Leakage | Pump |
| Gas Leakage | Accumulators |
| Switching char. degradation | Valve |

For the purpose of modeling the available is split into a training and a test dataset. To better reflect the reality of PHM use cases, where examples of relevant *Failure Incidents* are commonly extremely rare, we base our analysis on unsupervised learning methods that are fitted uniquely on healthy in-operation samples from the training dataset. This marks a distinction to Helwig's supervised learning approach, which relies on examples of Failure Incidents to train a fault classification model. Since our reference implementation's motivation lies in pointing out and consolidating a number of general best practices, we deliberately do not emphasize the selection of a specific ML algorithm. Without further model search or significant hyperparameter optimization, we use a standard similarity-based algorithm to fit a model, that yields a characteristic distance vector corresponding to each sample of the test dataset (Figure 3).

For the purpose of monitoring, this distance provides an *Attention Index*, indicative of how much a measured sample deviates from the healthy process. Introducing a threshold then yields potential failure detections. As visualized in Figure 4, the relationship between that threshold's value and the obtained precision (detections that are failures) and recall (failures that are detected) directly influences *TPR*, *FPR*, and *FNR*. Formula (4) suggests, there is an optimal threshold $x$ that maximizes $\Delta QCPN_{PHM}$:

$$\operatorname*{argmax}_{x} f(x, D, CD, CU, CDI, C_{PHM}) \tag{5}$$

This implies that determining the most cost-effective *Detection Method* not only demands a model well-adapted to the data expected in its specific environment and process parametrization (i.e. leading to a favorable precision-recall-curve). It also implies that its cost-effectiveness is highly dependent on its operational and maintenance context and an adequate threshold calibration must be aware of a failure mode's consequences, interventions, and their respective costs – information that is made available in the *DeepFMEA* data model.

To illustrate this, we evaluate the *Detection Method* in terms of its cost-effectiveness. We introduce two simplifications to Formula 4: (1) the default is a strict run-to-failure policy with no other mechanisms in place that are able to detect potential failures ($D \approx 0$); (2) the cost of operating the Detection Method is negligible compared to all over expected costs ($C_{PHM} \approx 0$). Lacking the concrete information about the cost of consequences and interventions in the reference use case, we formulate three characteristic scenarios, there corresponding $\Delta QCPN_{PHM}$-threshold-curves shown in Figure 5:

a) A non-critical *Failure Mode,* where the economic risk of tolerating failures is not significantly higher than managing them proactively ($CDI + CD \approx CU$). The *Detection Method* is never cost-effective.

b) A *Failure Mode*, where $CDI + CD < CU$. The *Detection Method* is cost-effective and displays a pronounced optimal threshold.

c) A critical *Failure Mode*, where the economic risk of failure significantly exceeds the cost of proactive failure management ($CDI + CD \ll CU$), i.e. due to high costs of downtime. The *Detection Method* is generally cost-effective.

Furthermore, its rich context in form of relationships between *System Elements, Virtual Sensors, Signals,* and *Failure Modes* permits us to enhance the prescriptive value of a *Detection Method's* outputs. Using the known semantic relationships within the system, we compute the extent to which the data points corresponding to a *System Element* contribute to the attention index and inform operators and maintainers which are the most likely *Failure Modes* for a detection. Despite the unsupervised nature of our approach, this achieves a "pseudo-classification" that provides insights at a diagnostic level. As indicated in Figure 6 on an exemplary basis, these can provide a high level of confidence given the highly distinguishable *Failure Modes* represented within the dataset.

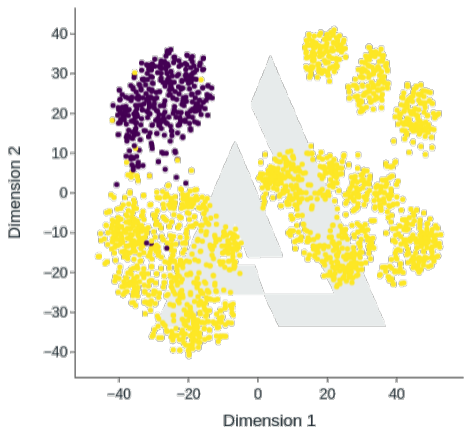

Figure 3. Two-dimensional projection of distance vectors yielded by an unsupervised hybrid model on the reference implementation's test dataset. Purple markers represent healthy cycles, yellow markers represent *Failure Incidents*.

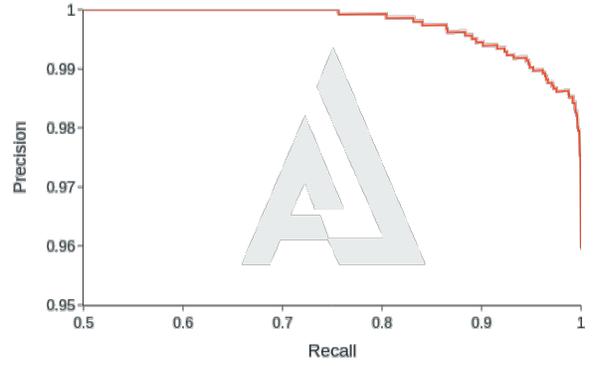

Figure 4. Precision-Recall Curve for the unsupervised monitoring algorithm on the hydraulic system dataset.

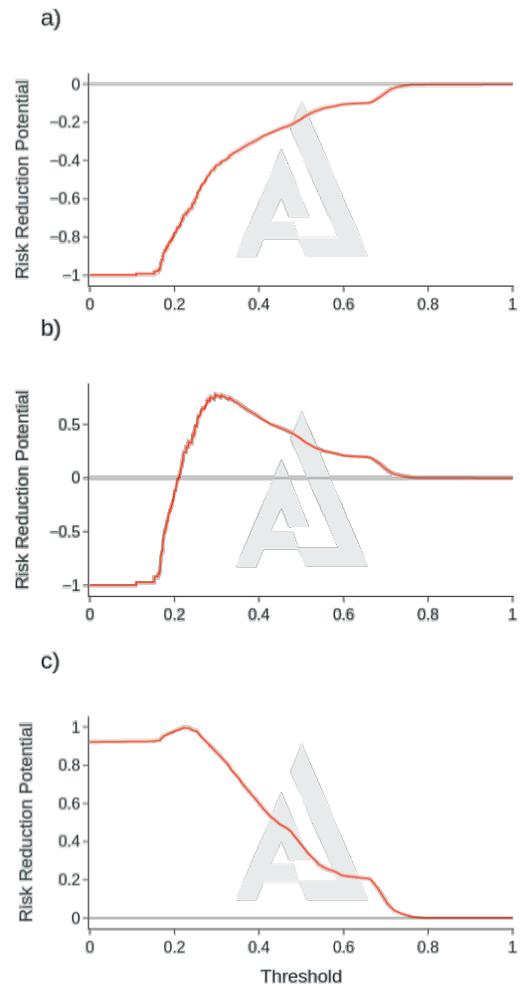

Figure 5. Expected Quantitative Consequence Priority Number reduction ($\Delta QCPN_{PHM}$) for the utilization of a *Detection Method*.

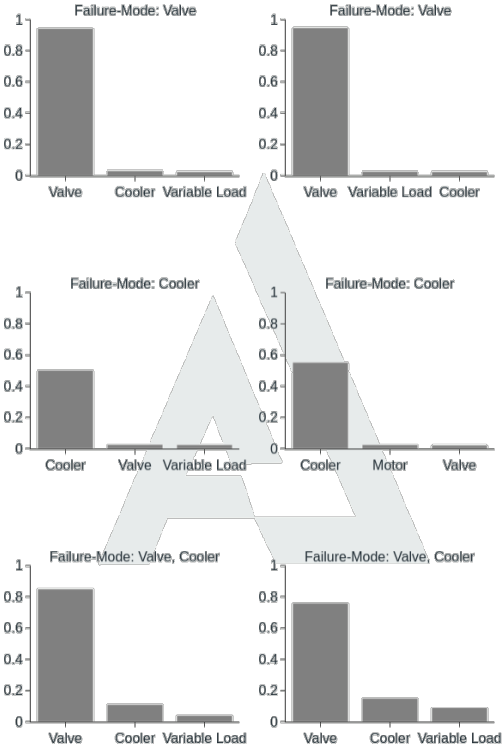

Figure 6 Top three contributions of *System Elements* to the attention index for 6 exemplary *Failure Instances*.

## 5. CONCLUSION: APPLICATION IN REAL-LIFE INDUSTRIAL SETTINGS

*DeepFMEA* introduces a novel abstraction to the design of data-driven PHM tools. It aims to:
- Enhance PHM development teams by incorporating best practices for systematically capturing and leveraging existing domain knowledge,
- Foster efficient collaboration among process experts, maintenance professionals, and data specialists in PHM projects,
- Streamline development workflows through standardization, which is essential for reusing specialized data management and MLOps modules. This standardization alleviates the burden of custom developments on PHM project budgets.

The application of this framework in real-world PHM projects across three equipment manufacturers has provided valuable insights:

- **Application-agnostic:** Tested in the steel, food & beverage, and machining industries, the framework's underlying data model has proven to be applicable for a broad range of similarly motivated use cases across different applications.
- **Model-agnostic:** Each implementation, whether employing simple dynamic thresholding techniques, classical ML algorithms and or advanced algorithms such as graph neural networks, benefited from *DeepFMEA's* abstraction of the detection method by enabling the use of reusable MLOps modules.
- **Extensible:** The framework accommodated various initial requirements, from simple monitoring to advanced diagnostics, offering clear pathways for evolving PHM tools to increase the prescriptive value of their outputs as more data becomes available.
- **Flexible:** It could be implemented both in new ("green field") projects and as an extension to existing data structures.
- **Automation:** Standardizing the capture and management of domain knowledge has facilitated the development of a graphical user interface, making PHM more accessible to non-data specialists, and reducing repetitive tasks for data specialists.

However, our experiences also highlight areas for improvement and expansion:
- **Risk Assessment Simplifications:** The current framework introduces strong simplifications for risk quantification and the assessment of the risk-reduction potential of PHM tools. It does not account for non-economic risks such as environmental and safety hazards. Moreover, it overlooks the multifaceted economic benefits of PHM, including waste reduction, energy efficiency, and productivity gains. A more refined model is necessary to accurately capture these aspects.
- **MLOps:** While *DeepFMEA* promotes standard MLOps practices for maintaining PHM tool trustworthiness, the unique challenges of PHM, such as limited benchmark data, organizational challenges in creating reliable data feedback loops to operators and maintainers, and varied operational contexts, demand more nuanced solutions. Addressing these challenges with innovative approaches should be a priority for the PHM research community and development teams to encourage broader adoption of data-driven PHM tools.